\documentclass{article}

\usepackage[final]{neurips_glfrontiers_2023}
\PassOptionsToPackage{numbers, compress}{natbib}

\usepackage{graphicx}
\usepackage{tikz}
\usepackage{amssymb}
\usepackage{hyperref}       % hyperlinks
\usepackage[english]{babel}

\usepackage{booktabs}
\usepackage{multirow}

\usepackage{amsmath}
\usepackage{mathtools}
\usepackage{amsthm}

\theoremstyle{plain}
\newtheorem{theorem}{Theorem}[section]
\newtheorem*{theorem*}{Theorem}

\newtheorem*{lemma*}{Lemma}

\theoremstyle{definition}
\newtheorem{definition}[theorem]{Definition}

\theoremstyle{remark}

\usepackage{amsfonts}
\usepackage{float}
\usepackage{array}
\usepackage{latexsym}
\DeclareMathOperator*{\argmax}{arg\,max}

\newcommand{\comp}{\mathcal{C}} %notation for connected components

\newcommand{\settitle}{\@maketitle}

\title{Graph Neural Networks on Discriminative \\Graphs of Words}

\author{%
  Yassine Abbahaddou \\
   LIX, École Polytechnique \\
   Institute Polytechnique de Paris, France  \\
  \texttt{yassine.abbahaddou@polytechnique.edu} \\
   \And
   Johannes F. Lutzeyer \\
   LIX, École Polytechnique \\
   Institute Polytechnique de Paris, France  \\
  \texttt{johannes.lutzeyer@polytechnique.edu} \\
   \AND
   Michalis Vazirgiannis \\
   LIX, École Polytechnique \\
   Institute Polytechnique de Paris, France  \\
   \texttt{mvazirg@lix.polytechnique.fr} \\
}

\begin{document}

\maketitle

\begin{abstract}
In light of the recent success of Graph Neural Networks (GNNs) and their ability to perform inference on complex data structures, many studies apply GNNs to the task of text classification. In most previous methods, a heterogeneous graph, containing both word and document nodes, is constructed using the entire corpus and a GNN is used to classify document nodes. In this work, we explore a new Discriminative Graph of Words Graph Neural Network (DGoW-GNN) approach encapsulating both a novel discriminative graph construction and model to classify text. In our graph construction, containing only word nodes and no document nodes, we split the training corpus into disconnected subgraphs according to their labels and weight edges by the pointwise mutual information of the represented words. Our graph construction, for which we provide theoretical motivation, allows us to reformulate the task of text classification as the task of walk classification. We also propose a new model for the graph-based classification of text, which combines a GNN and a sequence model. We evaluate our approach on seven benchmark datasets and find that it is outperformed by several state-of-the-art baseline models. We analyse reasons for this performance difference and hypothesise under which conditions it is likely to change. Our code is publicly available at: \href{https://github.com/abbahaddou/DGOW}{https://github.com/abbahaddou/DGOW}.

\end{abstract}

\section{Introduction}
Text classification is an important task in natural language processing. It has attracted  an increasing amount of attention following the success of Deep Learning. There are many application of text classification including sentiment analysis, intent detection and spam filtering \citep{DBLP:journals/corr/abs-1904-08067}. A key component of text classification is the representation of text. Recently,  an increasing body of work surrounding text classification suggests to model text corpora as graphs \citep{gamon-2006-graph,gokhan-etal-2022-gusum,phung-etal-2021-hierarchical}. The major benefit of graph representations is the ability to capture global information about the vocabulary, unlike sequential representations that are limited to local contextual information in sentences. Such approaches have two main components, the graph construction and the classification model. We will now introduce each of these in turn. 

Various graph constructions have been proposed to model text. In most cases, words and documents are represented by nodes in a graph and edges are drawn based on different relationship metrics, which we discuss now. Word co-occurrence is among the most popular relationship metrics; all the terms that co-occur within a fixed-size sliding window are linked by edges \citep{Rousseau2013, skianis-etal-2018-fusing, DBLP:journals/corr/abs-1809-05679}. Other works propose different weight computation for edges such as semantic \citep{Steyvers2005TheLS}, syntactic \citep{AlGhezi2020GraphbasedSW} and sequential relations \citep{10.1145/3041021.3055362,DBLP:journals/corr/abs-2001-05313}. In this work, we will also use a fixed-size sliding window to draw edges between word nodes in our Discriminative Graph of Words (DGoW) and weight edges by the pointwise mutual information of the represented words. The main distinguishing criterion of our construction is that we also take training labels into consideration in our graph construction, by constructing one disconnected subgraph per class. This construction allows us to forgo the use of document nodes and to better separate the information of the different classes as we demonstrate both theoretically and empirically.

Once a graph representation of the corpus is obtained, there are several approaches to the text classification task. 
There exist, non-deep-learning approaches \citep{skianis-etal-2018-fusing,gokhan-etal-2022-gusum} that first extract word and document embeddings from the graph structure, and then use machine learning algorithms, e.\,g., Support Vector Machines and Naive Bayes, to classify these embeddings. These methods outperform existing frequency-based criteria \citep{5392697,McCallum1998ACO}, but suffer from some limitations, such as high-dimensionality, data sparsity, and lack of flexibility. These two-step approaches, separating the graph embedding and classification model, can be simplified by the use of Graph Neural Networks (GNNs), which simultaneously perform the two steps, as has been done in many recent studies \citep{wu2019simplifying,zhang-etal-2020-every,DBLP:journals/corr/abs-2001-05313}. 
Some of these methods \citep{DBLP:journals/corr/abs-1809-05679,wu2019simplifying} operate in the \textit{transductive} learning setting, where both training and test sentences are used to construct the training graph.
While others \citep{Wang2022InducTGCNIG,DBLP:journals/corr/TangQM15} work in the \textit{inductive} learning setting, where only the training sentences are used in the graph construction. 
In this paper, we contribute the Discriminative Graph of Words Graph Neural Networks (DGow-GNN), in which we compose a GNN and a sequence model to simultaneously benefit from structural information of words in our DGoW construction and the order in which they arise. 

Our contributions can be summarised as follows.

\textbf{1)} We propose a \textit{new graph construction} by splitting the training corpus into disconnected subgraphs according to their labels and give theoretical motivation for our construction. This allows us to \textit{reformulate the problem of text classification as a walk classification task}. Formally, we predict the probability that a sentence is represented as walk in the subgraph of each class.

\textbf{2)} We propose a \textit{new model, the DGoW-GNN}, for graph-based representation of text which is a combination of a GNN and a sequence model. 

\textbf{3)} We perform \textit{extensive experimental validation} of our proposed graph construction and model on seven real-world benchmark datasets and observe that our DGoW-GNNs are outperformed by several state-of-the-art baseline models. We furthermore, analyse and explain these performance differences and hypothesise under which conditions they are likely to change.

% \begin{itemize}
%     \item[\textbf{1)}] We propose a \textit{new graph construction} by splitting the training corpus into disconnected subgraphs according to their labels and give theoretical motivation for our construction. This allows us to \textit{reformulate the problem of text classification as a walk classification task}. Formally, we predict the probability that a sentence is represented as walk in the subgraph of each class.
%     \item[\textbf{2)}] We propose a \textit{new model, the DGoW-GNN}, for graph-based representation of text which is a combination of a GNN and a sequence model. 
%     \item[\textbf{3)}] We perform \textit{extensive experimental validation} of our proposed graph construction and model on seven real-world benchmark datasets and observe that our DGoW-GNNs are outperformed by several state-of-the-art baseline models. We furthermore, analyse and explain these performance differences and hypothesise under which conditions they are likely to change. 
% \end{itemize}

\section{Related Work}
\label{sec:related_work}
We now introduce GNNs and give an overview of graph-based approaches to text classification.

\subsection{Graph Neural Networks}
Graph Neural Networks (GNNs) are neural networks that operate on graph-structured data, which is defined to be the combination of a graph structure, denoted by $G=(V, E),$ where $V$ and $E$ denote the vertex and edge sets, respectively, and a node feature matrix $X\in\mathbb{R}^{|V|\times d},$ containing the node feature vector of node $v_i$ in its $i^{\mathrm{th}}$ row.  Like most deep learning approaches, GNNs are formed by stacking several computational layers, each of which produce a hidden representation for each node in the graph, denoted by $H^{(\ell)} = [h^{(\ell)}_v]_{v\in V}$. A GNN layer $\ell$ updates node representations relying mainly (or only) on the structure of the graph and the output of the previous layer $H^{(\ell-1)}$. Conventionally, the node features are used as input to the first layer $H^{(0)} =X$. The most popular framework of GNNs is that of Message Passing Neural Networks \citep{Gilmer2017}, where the computations are split into two main steps:
% \begin{itemize}

    \textbf{Message-Passing}: Given a node $v$, this step applies a permutation-invariant function to its neighbours, denoted by $\mathcal{N}(v),$ to generate the aggregated representation, $$m^{(\ell)}_v= \text{\small AGGREGATE}^{(\ell)}(\{h^{(\ell-1)}_u, u \in \mathcal{N}(v)\}).$$ 
    
    \textbf{Update}: In this step, we combine the aggregated hidden states with the previous hidden representation of the central node $v,$ usually by making use of a learnable function, 
    $$h^{(\ell)}_v=\text{\small UPDATE}^{(\ell)}(h^{(\ell-1)}_v , m^{(\ell)}_v).$$
% \end{itemize}

Depending on the task, an additional readout or pooling function can be added after the last layer to aggregate the representation of nodes,  $$h_G = \text{ \small READOUT}(H^{(L)}).$$

Graph Convolutional Networks  \cite{DBLP:journals/corr/KipfW16} are among the most famous and commonly used GNN architectures. In GCNs, the graph convolutions are approximated by an order-one truncation of the expansion in terms of Chebyshev polynomials, which gives rise to a message-passing step in which weighted averages are taken over neighbourhoods in the graph. The weights in these weighted averages are fixed and depend on the square-rooted node degrees. A more general aggregation scheme is proposed in the more recent Graph Attention Networks \cite{veličković2018graph}, in which again a weighted average is used to combine information over graph neighbourhoods. The weights in these weighted averages are learned via a one hidden layer multi-layer perceptron taking both the central node's and neighbouring node's hidden states as input. Since the parameters of the attention mechanism in the GAT network were non-identifiable the Graph Attention Network V2 \cite{Brody2022HowAA} was proposed to yield a better-functioning attention mechanism, in which the weight matrices are separated by a non-linearity. Alternative standard GNNs include the Graph Isomorphism Network \cite{DBLP:journals/corr/abs-1810-00826}, which sums hidden states over neighbourhoods, and the GraphSage model \cite{DBLP:journals/corr/HamiltonYL17}, which uses a learned aggregator, in which the hidden states of the central node and neighbouring nodes are concatenated, processed by a learnable weight matrix and then aggregated using one of several proposed aggregation schemes. 
While our proposed DGoW-GNN can be defined on the basis of any GNN, without loss of generality, we make use of the GCN, the most commonly used GNN, in our proposed architecture here.

\subsection{GNNs For Text Classification}

\textit{TextGCN} \cite{DBLP:journals/corr/abs-1809-05679} was the first work to apply a GCN to text classification. The authors construct a heterogeneous graph with both word and document nodes, draw weighted word-document edges using TF-IDF weights and weighted word-word edges using the point-wise mutual information (PMI). The input graph in \textit{TextGCN} is constructed using both training and test documents, and a GCN is used to classify the document nodes. % of the constructed graph in a semi-supervised setting using the nodes representing the documents in training.  
 
Several works propose extensions of \textit{TextGCN}. Improvements are either made on the graph construction or the GNN architecture. For example, the authors of \textit{TensorGCN} \cite{DBLP:journals/corr/abs-2001-05313} combine three heterogeneous graphs which only differ in their word-word edge weights. One graph, relies on PMI weights, as is done in the \textit{TextGCN} approach, for the two other graph, semantic and syntactic based weights are used. The three graphs are fed to a GCN to perform text classification. Other models such as the \textit{HeteGCN} \cite{DBLP:journals/corr/abs-2008-12842} and \textit{SGCN} \cite{DBLP:journals/corr/abs-2106-05809} change the GNN architecture instead of the graph construction.

The methods discussed thus far in this section share one common problem: they are all transductive methods, i.e., the constructed graphs require both the training and the test documents. Thus, it is difficult to directly predict the labels of unseen documents. To deal with this issue, several works propose to work in the inductive learning setting. This is achieved in the \textit{TextING} \cite{zhang-etal-2020-every} and \textit{MPAD} \cite{nikolentzos2020message} models by constructing graphs on the sentence level, i.e., each sentence is represented as an individual graph. While the \textit{TextING} and \textit{MPAD} approach successfully implement a model capable of inductive learning, they have the drawback that each sentence graph only captures local information of the current sentence and global information, present in sentences not currently under consideration, is forgone. 
Subsequent inductive methods aimed to make use of the whole corpus by constructing training graphs from the entire training dataset. 
The \textit{InducT-GCN} \cite{Wang2022InducTGCNIG} for example generalises the \textit{TextGCN} by constructing the heterogeneous graph $G$ using only training sentences and training a GCN on it. To predict the label of an unseen small batch, the method creates a new graph $G'$ using both the training sentences and the batch. Due to the small size of the batch, the two graph $G$ and $G'$ are similar, so the GCN trained on $G$ can reasonably be expected to generalise to $G'.$ In another setting, \textit{PTE} \cite{DBLP:journals/corr/TangQM15} train a GNN on a graph constructed with only training documents, word embeddings are then extracted using the hidden representation of nodes in the GNN. To classify a new sentence, \textit{PTE} \cite{DBLP:journals/corr/TangQM15}  embed the sentence with the average of its words embeddings and then train small classifier (e.g, MLP) in an inductive setting.

Our novel graph construction and GNN-based model are applicable in the more realistic inductive learning settings, in which we do not have access to the test sentence structure during training.

\section{Graph Construction}
\label{graph_const}
\label{motivation}

\begin{figure*}[t]
    \centering
    \resizebox{\columnwidth}{!}{%
    \input{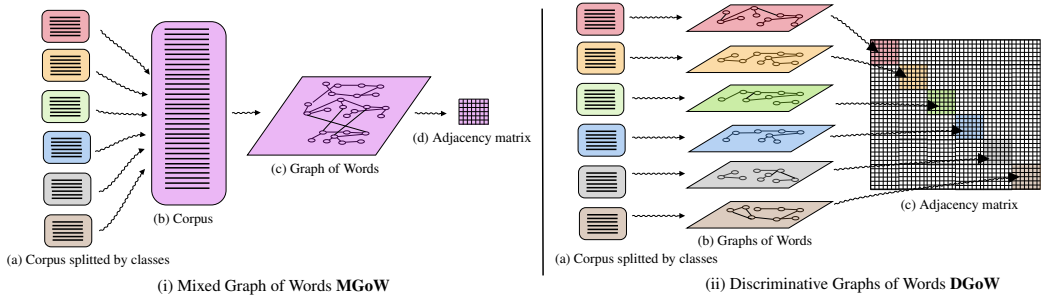}
    }
    \caption{Illustration of the two configurations MGoW and DGoW for a toy example of 6 classes in the dataset. The coloured boxes represent different classes. In the \textit{ MGoW}, we merge all the sentence into one corpus and construct one graph of words. In \textit{DGoW}, we keep the label based split, and create one disconnected subgraph per class.}
    \label{fig:GraphConstructions}
\end{figure*}

We begin by introducing our notation, formulating the problem of text classification and introducing necessary graph theoretical concepts. We then move on to present our proposed graph construction, the Discriminative Graph of Words (DGoW).  

\textbf{Notation.} Each training or test sentence $s$ is a sequence of words $s = [w^{(s)}_1,w^{(s)}_2,\ldots, w^{(s)}_{L(s)}],$ where $L(s)$ is the length of $s$. The number of words in a sentence depends on the data preprocessing used for the dataset. The label of a sentence belongs necessarily to a set of $P$ possible values. % : $\forall i, ~~y_i \in \{y_1,\ldots,y_P\}$. %We furthermore define a graph $G=(V,E)$ in terms of its node set $V$ and edge set $E.$

\textbf{Problem Formulation.} First of all, let us introduce the task of text classification. Given a training corpus, i.e., a set of training sentences  $\mathcal{S}^{train} = \{s^{train}_1, \ldots, s^{train}_N\}$ and their corresponding labels $\mathcal{Y}^{train} = \{y^{train}_1, \ldots, y^{train}_N\},$ the goal is to train a model to predict $\mathcal{Y}^{train}$  and generalise to the test corpus, i.e., the remaining unlabeled test sentences $\mathcal{S}^{test} = \{s^{test}_1, \ldots, s^{test}_M\}.$

We now define the graph theoretic concepts of walks, connected components and disconnected subgraphs in graphs, which will be central to our proposed graph construction. 

\begin{definition}[Walks, Connected Components and Disconnected Subgraphs]
A \textit{walk} in a graph $G=(V,E)$ is a sequence of vertices, such that any two vertices adjacent in the sequence are connected by an edge in $G.$ Then a subset $\mathcal{S}\in V$ is called a \textit{connected component} in $G$ if there exists a walk between any two vertices $v_i, v_j\in\mathcal{S}$ and no walk exists from any $v_i\in \mathcal{S}$ to any $v_j\notin\mathcal{S}.$ We further define \textit{disconnected subgraphs} $\comp_p$ for $p\in\{1,\ldots, P\}$ in a graph $G$ to be graphs whose node and edge sets partition the node and edge set of $G,$ respectively, such that there exists  no walk from any $v_i\in \comp_p$ to any $v_j\notin\comp_p$ for $p\in\{1,\ldots, P\}.$
\end{definition}

We furthermore weight edges in our proposed graph construction by the PMI of words, which is calculated as follows, 
$$ \text{PMI}(i,j) = \log\frac{p(i,j)}{p(i) ~ p(j)},$$
where $p(i,j) = \frac{\#W(i,j)}{\#W},$ $p(i) = \frac{\#W(i)}{\#W},$ $\#W$ denotes the total number of fixed-sized windows (i.e., fixed-sized word spans in the text), $\#W(i)$ denotes the number of windows containing the word $i$ and $\#W(i,j)$ equals the number of windows containing both words $i$~and~$j$ \citep{NIPS2013_9aa42b31}.

As discussed in Section \ref{sec:related_work}, the standard approach in the literature \citep{DBLP:journals/corr/abs-1809-05679,DBLP:journals/corr/abs-2001-05313,DBLP:journals/corr/abs-2106-05809} is to construct a graph on the basis of all sentences in the corpus, in which two words are linked based on different rules, which do not depend on the label of these sentences. These constructions typically contain both word and document nodes, where document nodes are linked to all words present in the represented documents. In the following we will refer to such graph constructions as \textit{Mixed Graphs of Words (MGoW)}. Since the text structure of a document is mainly encoded in the graph structure, representing the whole corpus in a single MGoW may reduce the capacity of GNNs to distinguish classes based on the structural characteristic of each class. We therefore propose an alternative graph construction. % in Definition \ref{def:DGoW}.

\begin{figure*}[t]
    \centering
    \resizebox{\columnwidth}{!}{%
    \input{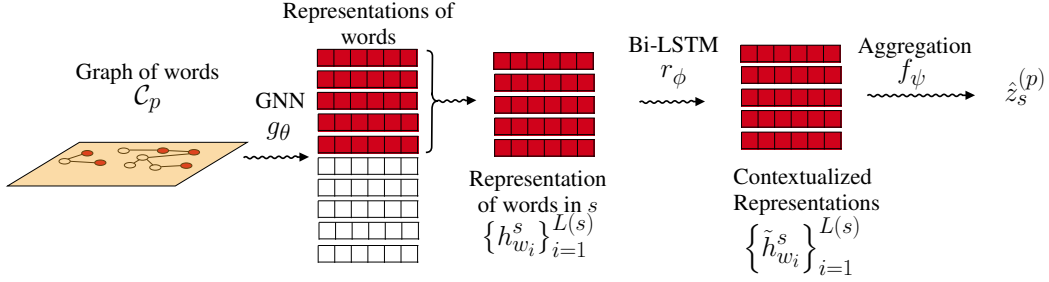}
    }
    \caption{The architecture of \textit{DGoW-GNN}. Our model takes as input the \textit{Discriminative} graph of words $\comp_p$, a class $p$ and a sentence $s$. The words occurring in $s$ are distinguished by the color red. As noticed, in the inductive setting,  the sentence representation in $\comp_p$ is not necessary a walk. We use the GNN $g_\theta$ to a vector representations of all the nodes in the graph. We select then only the vectors of the words occurring in $s$ and we fed it to a Bi-LSTM $r_\phi$ to contextualize the representations. At the end, we use a aggregation function $f_\psi$ to output a value in $[0,1].$}
    \label{fig:fig_model}
\end{figure*} 

\begin{definition}[Discriminiative Graph of Words] \label{def:DGoW}
In our Discriminiative Graphs of Words (DGoW) we construct one disconnected subgraph for each class in our training corpus. Edges in these disconnected subgraphs are drawn if words co-occur within a sliding window of size $\omega$ in the training corpus of the class corresponding to the given disconnected subgraph. Edges in our graph are weighted by the point-wise mutual information of the words represented by the connected word nodes. 
\end{definition}
Both the MGoW and DGoW constructions are illustrated in Figure \ref{fig:GraphConstructions}. Contrary to most baselines, only word nodes are considered in DGoWs. Our proposed model for document classification, to be introduced in Section \ref{proposed}, will not rely on document nodes to classify documents. 

In this work, we aim to stress the advantages of the DGoW construction over the MGoWs.  Formally, we want to show that, relying on graph structure, the DGoW construction brings sentences from the same class closer and separate sentences from different classes. Conceptually, separating the classes in the corpus into disconnected subgraphs, with no edges between nodes in different classes, separates the classes better and therefore should aid in the task of discriminating these classes. This intuition is formalised in the following theoretical result on the spectral node embeddings of graphs containing several connected components.

\begin{theorem} \citep{chung1997spectral}\label{thm:Spectral}
For a graph with $Q$ connected components the spectral node embeddings, produced by the normalised Laplacian eigenvectors corresponding to the smallest normalised Laplacian eigenvalue, are indicator vectors establishing the connected component membership of vertices.
\end{theorem}

We refer the reader to \citep[Proposition 4]{von2007tutorial} for the formal statement and proof of Theorem \ref{thm:Spectral}. Note that the disconnected subgraphs that correspond to the different classes may contain several connected components as there may exist training sentences that share no single word with the remaining sentences in their class. However, the sum of all eigenvectors indicating the connected components in a given disconnected subgraph is itself an eigenvector and hence, the eigenspace of the smallest normalised Laplacian eigenvalue can not only inidicate the connected components of a DGoW, but also the disconnected subgraphs it contains. Therefore, the spectral embeddings of nodes in different connected components will indicate the class membership of these nodes. In Section \ref{sec:StructuralEmbeddingResults} we will experimentally validate this theoretical insight for two different node embedding methods and clearly demonstrate that the classes are better separated in a DGoW than they are in a MGoW.

\section{Proposed Method}
\label{proposed}
\label{prop_method}

We now introduce our text classification model that takes DGoWs, construced as outlined in Section \ref{motivation} as input.
To take advantage of the structural differentiation in DGoW, we perform the classification of a sentence $s$ by evaluating the probability of it arising as a \textit{walk} of word nodes in each of the different disconnected subgraphs of our DGoW. This allows us to explicitly take the potentially different text structures of classes into account and thereby reformulate the task of text classification to correspond to the task of \textit{walk classification} in our DGoW. 
Thus, we train a neural network model $\mathcal{M}_\Theta$ to predict the probability that a sentence belongs to each disconnected subgraph. The predicted label of a sentence $s$ then corresponds to the class with the highest probability.
$$
    \hat{y}_s = \argmax_{q\in\{1,\ldots,P\}}\mathcal{M}_\Theta \left (  q|s, \{\comp_p\}^P_{p=1}  , \mathcal{S}^{train} \right ),
$$
where $\comp_p$ denotes the disconnected subgraph containing the sentence structure of class $p,$ $s$ is the input sentence and $\mathcal{S}^{train}$ is the set of sentences in the training set used in the graph construction, which does not necessarily include $s$ since we are working in the inductive learning setting. % For the transductive setting we use both training and test sentences, i.e., $\mathcal{S} = \mathcal{S}^{train}\bigcup \mathcal{S}_{test}$, whereas in the inductive setting we use only training sentences, i.e., $\mathcal{S} = \mathcal{S}^{train}.$ 

Our model $\mathcal{M}_\Theta$ is illustrated in Figure \ref{fig:fig_model}. It consists of three parts: the first part consist of a GNN $g_{\theta}$ to encode the nodes in graphs, the second part is a sequence model $r_{\phi}$ to capture the local context of words and the third part is an aggregation function $f_{\psi}$ to map the output into the desired format. This combination allows us to simultaneously capture the global contextual information of a word beyond the sentence currently under consideration with the GNN, as well as, the local information of the ordered words in their sentences. Below, we give further detail about each part of our model.

\textbf{Part 1 : Graph Neural Network. } Given a sentence $s=[w^{(s)}_1,w^{(s)}_2,\ldots , w^{(s)}_{L(s)}]$ belonging to a class $p$, we first encode the corresponding disconnected subgraph $\comp_p$ with a GNN $g_{\theta}$. The goal of this GNN is to refine the embedding of each word in the class $p$. Formally, we start by selecting the disconnected subgraph $\comp_p$ corresponding to the class $p$, we then feed it to $g_{\theta}$ and deduce the class dependent embedding of the words $\{w^{(s)}_i\}^{L(s)}_{i=1}$. To obtain the embedding of a word in all classes simultaneously, we can feed the entire DGoW to the GNN, where the adjacency matrix is block diagonal as illustrated in Figure \ref{fig:GraphConstructions}. 
We use $h^{(p)}_{w}$ to denote the GNN output for word $w$ in a chosen class $p$.
$$
    \left [ h^{(p)}_{w^{(s)}_1},h^{(p)}_{w^{(s)}_2},\ldots,h^{(p)}_{w^{(s)}_{L(s)}}\right ] = g_{\theta} \left (  s|\{\comp_p\}^P_{p=1}   \right ) .
$$
Without loss of generality, we use the GCN architecture as the GNN $g_{\theta}$ in our proposed model.

\textbf{Part 2 : Sequence Model.} We select the GNN embeddings of only the words occurring in the sentences $\{h^{(p)}_{w^{(s)}_i}\}_{i=1}^{L(s)}$. We feed the sequence of words into a sequence model $r_{\phi}$ , e.\,g.,  Bi-LSTM \citep{10.1162/neco.1997.9.8.1735}, to have contextualised embeddings and explicity benefit from the information contained in the ordering of the words in a sentence. 
$$
\left [  \Tilde{h}^{(p)}_{w^{(s)}_1},\ldots,\Tilde{h}^{(p)}_{w^{(s)}_{L(s)}}\right ] = r_{\phi}\left ( \left [ h^{(p)}_{w^{(s)}_1},\ldots,h^{(p)}_{w^{(s)}_{L(s)}}\right ] \right ).
$$

\textbf{Part 3 : Aggregator.} Now that the embedding of each word depends on both the class structure and the context of the sentence, we aggregate the embedding using a function $f_{\psi}$. In our case, we simply average each output of the Bi-LSTM and feed the new representation to a Multi-Layer Perceptron (MLP) to produce a predicted probability value $\hat{z}^{(p)}_s \in [0,1]$ for each class $p =1,\ldots, P.$
$$
 \hat{z}^{(p)}_s= f_{\psi}\left ( \left [  \Tilde{h}^{(p)}_{w^{(s)}_1},\ldots,\Tilde{h}^{(p)}_{w^{(s)}_{L(s)}}\right ] \right ).
$$
We next predict sentences to belong to the class with the highest predicted probability, i.e.,
$$
\hat{y}_s = \argmax_{q\in\{1,\ldots,P\}}  \hat{z}^{(p)}_s .
$$

To train our model to perform these individual binary predictions of how likely a given sentence is to arise in a given class $p$ we have to make use of negative samples during training. We use the term negative samples for a class $p$ to describe sentences, which are sampled from the corpus excluding $p.$ It will be the task of our model to predict that these negative samples do not belong to the currently considered class $p.$ Specifically, in our training procedure each sentence is considered twice. Firstly, we look at the sentence in its corresponding class, i.\,e., $y_i = p$, in this case, we train $\mathcal{M}_\Theta$ to predict the value of 1 for  $\hat{z}^{(p)}_s $. Secondly to obtain negative samples, we randomly select a class $q$ different form the class label, i.\,e., $y_i\neq q,$ and we train $\mathcal{M}_\Theta$ to predict the value of 0 for $\hat{z}^{(q)}_s $. Since, we perform the task of binary classification, we use the Binary Cross-Entropy loss. %\todo{add detail on why this loss is necessary, since Giannis asked about it: 1) Arises naturally as a consequence of our well-separated graph construction 2) test components separately}
% $$
%     \mathcal{L}(s,p) = \left\{
%     \begin{array}{ll}
%         \log(\hat{z}^{(p)}_s ), & \mbox{if } y_i=p; \\
%         \log(1-\hat{z}^{(p)}_s ), & \mbox{otherwise.} 
%     \end{array}
% \right.$$

% We give additional details about the model architectures, the training and test setup in Section \ref{expreim_setup}. We run the the experiments  for inductive and transductive settings. We present the results in Section \ref{results}.

\section{Experiments and Results}
% % \section{Experimental Setup}
\label{expreim_setup}
In this section, we present the benchmark datasets used for our experiments, the selected baselines and the process used for the training and evaluation. 
We furthermore present experiments in which we observe the DGoW construction to lead to better class separation than a MGoW in Section \ref{sec:StructuralEmbeddingResults}. Finally we present the results of our DGoW-GNN and our baseline models on seven real-world benchmark datasets in Section \ref{sec:RealWorldExperiments}, as well as an analysis explaining the performance of our model together with several ablation studies highlighting the impact of different model choices in our DGoW-GNN architecture in Section \ref{sec:ablation}. 

We provide information about implementation details of our experiments, including optimal hyperparameters, in Appendix \ref{app:ImplementationDetails}. All source code is publicly available on GitHub \footnote{Code available at \href{https://github.com/abbahaddou/DGOW}{https://github.com/abbahaddou/DGOW} }. 
    
\textbf{Datasets.} 
For a fair comparison, we use five datasets used throughout the graph-based text classification literature \citep{DBLP:journals/corr/abs-1809-05679,DBLP:journals/corr/abs-2001-05313,zhang-etal-2020-every,DBLP:journals/corr/abs-2112-06386,ding-etal-2020-less}. In particular we run experiments on the \textit{Reuters 8 (R8)} and \textit{Reuters 52 (R52)} \citep{10.1145/183422.183423}, \textit{Ohsumed (OH)} \citep{hersh1994ohsumed}, \textit{Movie Review (MR)} \citep{pang-lee-2005-seeing} and  \textit{20 Newsgroups (20NG)} \citep{LANG1995331} datasets. In addition, we include the \textit{BBC News (BBC)} \citep{greene06icml} and \textit{Internet Movie Database (IMDb)} \citep{maas-EtAl:2011:ACL-HLT2011} datasets to test our model on long documents as well as very large datasets. In Appendix \ref{app:datasets} we provide further details on these datasets and their summary statistics, as well as further information on our data pre-processing. 

\textbf{Baselines.} 
Since our DGoW-GNNs apply in the inductive learning setting, we benchmark the performance of our model against the state-of-the art graph-based inductive methods. 
In particular, we consider the \textit{InductTGCN}  \citep{Wang2022InducTGCNIG}, \textit{TextING}  \citep{zhang-etal-2020-every} and \textit{HyperGAT} \citep{ding-etal-2020-less}  for experimental comparison.

\subsection{Structural Embedding Experiments}
\label{sec:StructuralEmbeddingResults}

\begin{table}[t]

\centering
% \resizebox{0.6\columnwidth}{!}{%
% \scriptsize
\caption{Structural similarity, measured via spectral node embeddings, between different pairs of labels for the R8 dataset. %, \textcircled{1} MGoW \textcircled{2} DGoW.
}\label{tab:R8sim}
\begin{tabular}{lr|rrrrrr}\toprule
&  \textbf{Labels}  & $\mathbf{\boldsymbol{\omega}=2}$ & $\mathbf{\boldsymbol{\omega}=5}$  & $\mathbf{\boldsymbol{\omega}=10}$  & $\mathbf{\boldsymbol{\omega}=15}$& $\mathbf{\boldsymbol{\omega}=20}$\\ \cmidrule{1-7}
\multirow{3}{*}{MGoW} 
&\textit{earn/earn}  & 84.68 & 70.18  & 67.23 & 65.29 & 66.96 \\
&\textit{acq/acq}    & 87.93 & 68.57 & 62.47 & 58.90& 57.97\\ 
&\textit{earn/acq} & 80.67 &  36.78 &  26.74 & 25.75 &  26.18\\ \cmidrule{1-7}

\multirow{3}{*}{DGoW} 
&\textit{earn/earn} & 90.26 & 80.90 & 71.36 & 72.46 & 72.06 \\
&\textit{acq/acq} & 99.58 & 89.03 &  87.75 & 84.21 & 83.19\\
&\textit{earn/acq} & 0 &  0 & 0  & 0 & 0\\
\midrule
\end{tabular}
% }
\end{table}

We now present a set of experiments which aims to measure the structural separateness of sentences in different classes in both the MGoW and DGoW constructions. To do so, we obtain node, i.e., word, embeddings in the two graph constructions using either the \textit{spectral embeddings} obtained from the symmetrically normalised graph Laplacian \citep{chung1997spectral,DBLP:journals/corr/abs-1809-11115} or the \textit{FastGAE} model \citep{10.1016/j.neunet.2021.04.015}. We then represent sentences as the sum of their respective word embeddings. To measure the structural similiarity of classes in the graph we make use of the cosine similarity of sentence embeddings,  
$$\delta(p,q, h)  =  \frac{1}{|\comp_p|  |\comp_q|} \sum_{i \in \comp_p}\sum_{j \in \comp_q; ~i\neq j} \frac{ h_i^\top h_j}{\lVert h_i\lVert\lVert h_j\lVert},$$
where $h_i$ and $h_j$ denote sentence embeddings and  $\lVert\cdot\lVert$ is the $L_2$ norm of vectors.
When $p \neq q$, $\delta$ measures the intra-similarity between the two classes, i.\,e.,  to which degree the subgraphs, representing two different sentences of labels $p$ and $q$, are structurally similar.  On the other hand, when $p=q$, $\delta$ measures the inter-similarity between sentence embeddings within the same class. In this experiment we only consider classes with more than 400 sentences to get reasonably stable estimates. After constructing the graph on the basis of the whole corpus, we furthermore, randomly sample 400 sentences in each class to compute the two similarity metrics on the basis of these.

We report the result for the R8 dataset in Table \ref{tab:R8sim}. We notice the structural similarity of sentences in the MGoW configuration  to be very high even when their two classes are different. On the other hand, the similarity between different classes in the DGoW configuration is almost null for different classes. In other words, while the structural inter-similarity is high in both configurations, the correlation between the graph representations of two sentences from two different classes is almost null in DGoW. We can benefit from the structural differences in DGoW to easily distinguish the classes. 
In Appendix \ref{app:embedding} we show results obtained on the OH dataset and also for the \textit{FastGAE} \citep{10.1016/j.neunet.2021.04.015} embedding method run on both the R8 and OH datasets. These additional experiments all support the conclusions drawn on the basis of Table~\ref{tab:R8sim}.

Another conclusion we can draw concerns the effect of the window size in Table \ref{tab:R8sim}. As the window size increases, the similarities decrease. So, increasing the window size helps differentiate different classes, but has the counter-effect on sentences belonging to the same class. Since, the intra-similarity is almost constant in the DGoW configuration, the best window size is $\omega=2$.

\begin{table*}[t]\centering
% \tiny
% \scriptsize
\caption{Results of different models on the benchmark datasets; \textcircled{1} Inductive approaches \textcircled{2} Sequence models. %For the baseline \textit{TensorGCN}, we run the official code of the paper\footnote{\href{https://github.com/THUMLP/TensorGCN_pytorch}{https://github.com/THUMLP/TensorGCN\_pytorch}}, we didn't manage to reproduce the result mentioned in the paper, so we directly report the result published in the paper.
}\label{tab:PGM}
\resizebox{1\columnwidth}{!}{%
\begin{tabular}{lllllllllll}\toprule
&\textbf{Model} &\textbf{Reference}  & \textbf{R8} & \textbf{R52}  & \textbf{OH}  & \textbf{MR} & \textbf{20NG} & \textbf{BBC} & \textbf{IMDB}\\ \cmidrule{1-10}
\multirow{3}{*}{\textcircled{1}} 
&InductTGCN & \citep{Wang2022InducTGCNIG} & \textcolor{black}{96.60 (0.17)} & \textcolor{black}{93.18 (0.23)} & \textcolor{black}{66.23 (0.48)} & \textcolor{black}{75.75 (0.50)} & \textcolor{black}{90.64 (0.21)} & \textcolor{black}{96.36 (0.18)}  & \textcolor{black}{86.36 (0.29)}\\
&TextING & \citep{zhang-etal-2020-every} & \textcolor{black}{97.19 (0.30)} & \textcolor{black}{94.24 (0.30)} & \textbf{\textcolor{black}{69.55 (0.45)}} & \textcolor{black}{79.40 (0.44)} & - & \textcolor{black}{97.54 (0.14)} & \textcolor{black}{-} \\
&HyperGAT & \citep{ding-etal-2020-less} & \textcolor{black}{96.53 (0.25)}  & \textcolor{black}{92.75 (0.28)} & \textcolor{black}{ 62.64 (0.61)} & \textcolor{black}{76.76 (0.31)} & \textbf{\textcolor{black}{91.34 (0.17)}} & \textcolor{black}{96.56 (0.33)} & \textcolor{black}{86.25 (0/12)} \\

 & \textcolor{black}{DGoW-GNN w/o Bi-LSTM} & Ours  & \textcolor{black}{94.11 (1.12)} & \textcolor{black}{82.98 (1.19)} & \textcolor{black}{30.01 (2.30)} & \textcolor{black}{67.89 (0.41)} & \textcolor{black}{71.00 (1.75)}& \textcolor{black}{89.64 (0.53)}  & \textcolor{black}{68.66 (0.56)} \\
& DGoW-GNN & Ours& \textcolor{black}{95.17 (0.22)} & \textcolor{black}{86.26 (1.54)} & \textcolor{black}{44.59 (1.01)} & \textcolor{black}{71.65 (0.39)} & \textcolor{black}{79.21 (1.29)} & \textcolor{black}{92.14 (6.32)} & \textcolor{black}{76.76 (1.90)} \\
\cmidrule{1-10}
\multirow{2}{*}{\textcircled{2}} 
&Bi-LSTM & & \textcolor{black}{94.41 (0.60)} & \textcolor{black}{86.09 (2.15)} & \textcolor{black}{36.90  (1.11)}& \textcolor{black}{72.36 (0.86)} & \textcolor{black}{71.24 (1.25)} & \textcolor{black}{86.96 (3.05)} & \textcolor{black}{86.38 (0.37)} \\
&BERT  & \citep{DBLP:journals/corr/abs-1810-04805} & \textcolor{black}{97.78 (0.20)} & \textcolor{black}{93.21 (0.21)} & \textcolor{black}{61.86 (0.72)} & 85.30 (0.31) &  \textcolor{black}{86.22 (0.31)} & \textcolor{black}{97.77 (0.33)} & \textcolor{black}{70.53 (0.72)} \\
&RGCN-BERT  & Ours & \textbf{98.01 (0.61)} & \textbf{93.58 (0.57)} & 62.41 (0.81)  &   \textbf{86.13 (0.54)} & 87.43 (0.67) & \textbf{98.01 (0.41)}  & \textbf{87.02  (0.64) }\\
\bottomrule
\end{tabular}
}
\end{table*}

% \section{Results}
% \label{results}
% \subsection{Similarity Between The Structural Embedding }

\subsection{Results of DGoW-GNN}
\label{sec:RealWorldExperiments}

\begin{figure}[t]
    \centering
    \includegraphics[width=0.8\columnwidth]{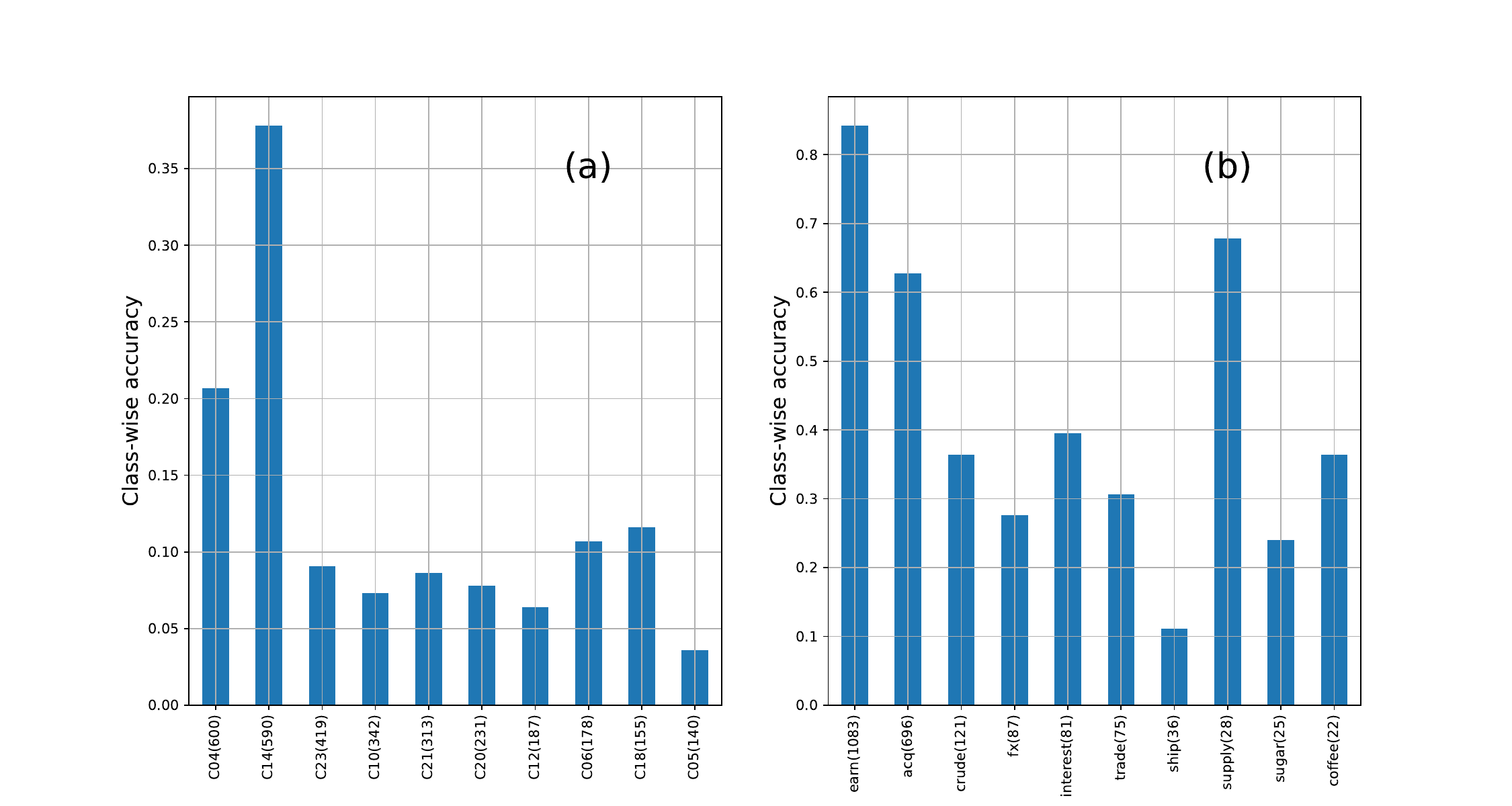}
    \caption{Class-wise accuracy of the DGoW-GNN on the OH (a) and R8 (b) datasets.}
    \label{fig:ClassWiseAccuracy}
\end{figure}

We now analyse the results of \textit{DGoW-GNN} in the inductive learning setting. We compare to graph-based approaches as well as the Bi-LSTM, a finetuned BERT and a combination of the BERT and RGCN  model \citep{Schlichtkrull2018}. We report all the results in Table \ref{tab:PGM}.

We observe that our \textit{DGoW-GNN} is outperformed by the graph-based baseline models. We believe this performance difference to arise as a result of insufficient context being accessible within our graph construction in the perfectly separated disconnected subgraphs. Indeed, we verify this hypothesis for the R8 and OH datasets in Figure \ref{fig:ClassWiseAccuracy}, where we are clearly able to observe that the performance of our  \textit{DGoW-GNN} on sentences in the different classes correlates with the number of training sentences in these classes. We therefore hypothesise that our DGoW graph construction and \textit{DGoW-GNN} has the potential to outperform the state-of-the-art baselines on larger datasets, in which we have more training sentences per class to add sufficient context to each word node.  

We furthermore notice that our \textit{DGoW-GNN} consistently outperforms the \textit{Bi-LSTM}, which highlights the positive contribution of the graph construction to the model performance. We observe the graph-based models, including the \textit{DGoW-GNN}, to outperform  BERT on the IMDB and 20NG datasets, since the graph-based approaches are better adapted to long documents than BERT which, due to the high complexity, uses truncation and thus loses crucial information from the documents. The superiority of BERT on the remaining datasets can be explained by the power of pretrained models on short sentences \citep{joshi2019bert}. 

We investigate the RGCN-BERT model, in which we work with a MGoW, where edges are typed according to the training class from which they originate. This allows us to make use of the RGCN \citep{Schlichtkrull2018}, in which we aggregate and update over the different types of edges separately and then aggregate the type-wise representation. The node embeddings from the RGCN are then concatenated with the corresponding word embeddings from the BERT model to be fed to final classifier. We clearly observe that this heterogeneous graph construction falling in between the MGoW and DGoW construction in conjunction with the BERT embeddings outperforms all other baselines.

\subsection{Ablation Studies}
\label{sec:ablation}

For the graph construction, we obtain the best results with a window sizes $\omega=2$ as observed in Appendix \ref{sec:WindowSizeAblation}. As in previous work \citep{kiela2014systematic,keselj2009speech},  smaller window sizes help produce syntactic representations of words and also capture relations other than co-occurrence, e.\,g., dependency relations.

We further study the effect of different word embeddings as node features in our DGoW by training our \textit{DGoW-GNN} also on DGoWs with the pre-trained GLoVe embeddings as node features \citep{pennington2014glove}. We provide the experimental results in Appendix \ref{sec:WordEmbeddingAblation}. In most datasets, we obtain better results using the one-hot encodings. As noticed by \citep{galke2022bag}, \textit{pretrained embeddings, e. g., GloVe, can have a detrimental effect on model performance.} Therefore, we use one-hot encodings of the represented words as node features in our DGoW for our \textit{DGoW-GNN} and all baseline models.

To study the importance of combining a GCN and a Bi-LSTM in \textit{DGoW-GNN}, we train only the GCN and the aggregation function, and we omit the Bi-LSTM. We refer to this model as \textit{DGoW w/o Bi-LSTM}. We used the same graph construction and the same training setup. We also trained separately a Bi-LSTM on the multi-label classification task. We report the comparison results in Table~\ref{tab:PGM}.  We notice an improvement in the classification accuracy when mixing the GCN and the Bi-LSTM in our DGoW-GNN model.

We also test two different aggregation functions (\textit{MLP} and $\text{\small PROD}$), in addition to the average aggregation function $\text{\small AVG}$. These aggregation functions are described in Appendix~\ref{sec:AggregatorAblation} together with the result from this ablation study. We clearly observe the best results using the $\text{\small AVG}$ aggregator.

The extensive ablation studies in this section show that we present a well-optimised and sufficiently explored approach in our DGoW-GNNs.

\section{Conclusion}
In this work, we present a new graph construction, the DGoW, for the task of text classification. We show both theoretically and in practice that our DGoW better separates the classes to be recovered in text classification.  
We also propose a new graph-based model \textit{DGoW-GNN}, which is a combination of a GNN, a sequence model, and an aggregation function. 
Our experiments demonstrate that \textit{DGoW-GNN} is outperformed by state-of-the-art graph-based approaches for text classification in the inductive learning setting.  
While our \textit{DGoW-GNN} does not outperform the existing state-of-the art baselines, we believe it to be a well-motivated idea, which furthers our understanding of the graph-based approach to text classification and has the potential to lead to performance improvements at large scale. 

%We therefore argue that our work makes an interesting contribution to the literature, worthy publication to advance scientific research in this direction. 
 % As explained, on some datasets \textit{DGoW-GNN} may fail if some labels are not enough frequent in the training dataset. 

 %As future work, we would like to overcome this limitation by proposing an extension model of  \textit{DGoW-GNN} that combines the Mixed Graph \textbf{MG} and the Discriminative Graph \textbf{DG} in the inductive setting. The new model will add knowledge for class labels in \textbf{DG} that are not frequent in the training dataset. This combination might be at the graph construction level or in the GNN architecture. Another interesting direction of this work is to combine \textit{DGoW-GNN} with the state-of-the-art model \textit{BERT}. As a starting point, we could constract the graph of words using the same data processing of BERT,i.\,e. Fast wordpiece tokenization \citep{song2020fast}, and then use a fusion strategy between a GNN and BERT.

\section*{Acknowledgements}
This work was partially supported by ANR via the AML-HELAS (ANR-19- CHIA-0020) project. The computation (on GPUs) was performed using HPC resources from GENCI-IDRIS (Grant 2023-AD010613410R1).

% % \newpage

% \bibliographystyle{plain}
\bibliography{references}
\bibliographystyle{unsrtnat}

\newpage

\appendix

\vbox{%
\hsize\textwidth
\linewidth\hsize
\vskip 0.1in
% \@toptitlebar
\centering
{\LARGE\bf Supplementary Material: Graph Neural Networks on Discriminative Graphs of Words\par}
% \bottomtitlebar
\vspace{2\baselineskip}
}

% \begin{center}
%     \textbf{Supplementary Material: Graph Convolutional Networks With Orthogonal Weights Are More Robust}
% \end{center}

\section{Implementation Details}
\label{app:ImplementationDetails}
To generate the DGoW, we used a fixed-sized sliding window of size 2. We fed the adjacency matrix of the DGoW into a Graph Convolution Network (GCN). We used the identity matrix as node features for the GCN. The number of GCN layers depends on the dataset, e.\,g. we use 2 layers for MR, 20NG and IMDB, 3 layers for R8, R52 and OH and 4 layers for BBC. For the aggregator, we use a Multi-Layer Perceptron with one hidden layer of dimension 128 and a scalar output. We use the ReLU activation function after the first layer and the sigmoid function after the second layer to output a value in the range 0 to 1.

To generate a random different label during training, we use \textit{Frequency based negative sampling}, i.\,e., we sample labels according to their frequency in training. More precisely, given a training sentence from the class $\comp_p$, we randomly sample a new label using the weights
$ \tfrac{|\comp_q|}{\sum_s |\comp_s|}$ for $ q\neq p.$ 

We train our model on the seven datasets in the inductive settings using the Adam optimiser with a learning of $10^{-3}$. We repeated the training 10 times to test the stability of the model.  To avoid overfitting, we randomly consider 10\% of training sentences as validation set. In the inductive setting, we do not include the validation sentences in the graph construction. We stops the training as soon as the validation accuracy becomes worse than the four previous values.

To generate the structural embedding of \textit{FastGAE}, we use 2-Layer \textit{GCN} as the GAE autoencoder and we choose an embedding size of 256. We train it on Binary cross-entropy reconstruction loss using Adam optimiser \cite{Kingma2014} with a learning rate of $10^{-2}$. 

For the spectral embedding, we chose 16 as the dimension of our embeddings. Since our embedding corresponds to the eigenvectors of the normalized Laplacian matrix and these eigenvectors are indicator vectors establishing which vertex is an element of which connected component \cite{von2007tutorial}, so the dimension should be greater than the number of connected components in our DGoW.

We also include sequence models in our baselines. We choose to train a BI-LSTM and we also finetune a pre-trained BERT model on the multi-label classification task. We report the result of all these baselines with the results of our model \textit{DGNN} in Table \ref{tab:PGM}.

To produce the baseline results, we used the code provided by the authors of %\textit{TextGCN}\footnote{\href{https://github.com/yao8839836/text_gcn}{https://github.com/yao8839836/text\_gcn}}, \textit{TensorGCN}\footnote{\href{https://github.com/THUMLP/TensorGCN_pytorch}{https://github.com/THUMLP/TensorGCN\_pytorch}}, 
\textit{InductTGCN}\footnote{\href{https://github.com/usydnlp/inducttgcn}{https://github.com/usydnlp/inducttgcn}}, \textit{TextING}\footnote{\href{https://github.com/CRIPAC-DIG/TextING}{https://github.com/CRIPAC-DIG/TextING}} and \textit{HyperGAT}\footnote{\href{https://github.com/kaize0409/HyperGAT_TextClassification}{https://github.com/kaize0409/HyperGAT\_TextClassification}}. We made sure that all code used the same data processing which will be described in Appendix \ref{app:embedding}. We trained the baselines 10 times to compare the stability of each model. We were not able to produce the result of \textit{TextING} for the 20NG dataset due to the high memory consumption of the model for high datasets, even by using RTX A6000 GPU. 20NG was also the only dataset in the original paper where the authors didn't report the results.  For the model \textit{TextING}, we didn't find any implementations, so we directly took the results from the original paper.  %Also for the baseline \textit{HeteGCN}, we didn't find an official implementation, so we directly reported the result mentioned in the paper.

\section{Further Information on The Considered Datasets}
\label{app:datasets}
For a fair comparison, we used five datasets used in the graph-based text classification papers \cite{DBLP:journals/corr/abs-1809-05679,DBLP:journals/corr/abs-2001-05313,zhang-etal-2020-every,DBLP:journals/corr/abs-2112-06386,ding-etal-2020-less}. Below, we give a short description on the used datasets.

 \textbf{Reuters 8 (R8)} and \textbf{Reuters 52 (R52)} \cite{10.1145/183422.183423}: two datasets collected from the Reuters financial newswire service. The classes represent topics such as business, sports, science, and technology. 

 \textbf{Ohsumed (OH)} \cite{hersh1994ohsumed}: medical information database, consisting of titles and abstracts from medical journals. 

 \textbf{Movie Review (MR)} \cite{pang-lee-2005-seeing}: a binary sentiment classification dataset containing movie reviews.

 \textbf{20 Newsgroups (20NG)} \cite{LANG1995331}: a  dataset comprising around 18,000 newsgroup posts on 20 different newsgroups.  The newsgroups cover a wide range of topics, including computers, politics, and sports.

In addition to these datasets, we include IMDB and BBC datasets to test our model on long documents as well as very large datasets.

 \textbf{BBC News (BBC)} \cite{greene06icml}:  a dataset of news articles from the BBC news website, labeled with one of five categories: business, entertainment, politics, sport, and technology.

 \textbf{Internet Movie Database (IMDB)} \cite{maas-EtAl:2011:ACL-HLT2011}: a dataset of 50,000 movie reviews labeled as either positive or negative. The dataset is balanced, meaning that 
there are an equal number of positive and negative reviews. 

\begin{table}[t]
\centering
\caption{Basic statistics of the benchmark datasets.}
\label{tab:table_dataset}

\begin{tabular}{l|rrrrrr}\toprule
\textbf{Dataset} & \textbf{\# Docs} & \textbf{\# Train} & \textbf{\# Test} & \textbf{\# Class} & \textbf{\# Vocabulary} & \textbf{Avg. Length} \\ \hline
R8      & 7,674 & 5,485 & 2,189 &  8 &  12,150 &  67         \\
R52     & 9,100 & 6,532 & 2,568 & 52 &  13,769 &  71  \\
OH      & 7,400 & 3,357 & 4,043 & 23 &  12,258 &  133    \\
MR      & 10,662& 7,108 & 3,554 & 2  &  7,869  &  19     \\
20NG    &  18,846 &  11,314 & 7,532 & 20 & 40,852 &  220  \\
BBC     &  2,225  &  1,225 & 1,000 & 5 & 14,380 & 217  \\
IMDB    &  50,000  & 25,000  & 25,000 & 2  & 48,837 & 141   \\\hline
\end{tabular}
\end{table}
In Table \ref{tab:table_dataset}, we present some basic statistics of the used datasets. More precisely, we present the total number of sentences, the number of training and test sentences, and number of classes. We also give the size of vocabulary and the average length of sentences in each dataset after the tokenisation. We notice the type of datasets is varied; we have datasets, such as IMDB, with a very large number of sentences, as well as  very small datasets such as BBC.  We also test datasets with varied vocabulary size and average length of documents.

We follow the same data processing used in baselines. Formally, except for the MR dataset, we remove non-alpha numerical characters, the leading and the trailing characters. All characters are converted to lowercase. And we finally split a sentence into words with a white-space character. In Table \ref{tab:table_dataset}, we give supplementary information about the datasets. Additionally, we also remove all words that occur less than 2 times in the training data, and remove stop words (except for MR dataset as they improve the performance of models in the sentiment analysis task). There exist other tokenisation techniques in NLP  \cite{chen2015convolutional, sennrich2015neural}, but for consistency with previous work on graph-based approaches for text classification and, without loss of generality, we choose to adopt the same tokenisation as \cite{Wang2022InducTGCNIG} for our method and for all used baselines. 

% We choose to adopt the same tokenisation method as previous studies.

\section{Additional Structural Embedding Results}
\label{app:embedding}

To support our DGoW configuration, we generated the structural embedding of words in both DGoW and MGoW configuration. Since we have multiple graphs in DGoW, we need to generate embeddings for each graph and it is important to generate all the embedding in the same vector space. To do so, we use the DGoW containing the several disconnected subgraphs  illustrated in Figure \ref{fig:GraphConstructions} and do not consider disconnected subgraphs in isolation.  As mentioned in Section \ref{motivation}, we used different node embedding methods : \textit{Spectral Embeddings} \cite{chung1997spectral,DBLP:journals/corr/abs-1809-11115} and \textit{FastGAE Embeddings} \cite{10.1016/j.neunet.2021.04.015}.
We generate the embeddings on two different dataset OH and R8. Since the number of nodes in the DGoW can be as large as $|\mathcal{V}| \times P$ where $P$ is the number of classes and $\mathcal{V}$ the vocabulary size, we keep only the 4 most frequent classes in OH when generating spectral embeddings as the methods is time-consuming with very high computational complexity. For a fair comparison for OH, we also keep the  4 most frequent classes in the MGoW configurations. For the embeddings of  R8 and for the embeddings of OH when using \textit{FastGAE}, we keep all the classes.

\subsection{Results Structural Similarity Using Spectral Embedding}

\label{sec:appendix_A}

In Table \ref{tab:OHsim}, we report the result of the experiment comparing the \textit{intra-similarity} and \textit{inter-similarity} of the spectral embedding for the dataset OH. As discussed in  Section \ref{expreim_setup}, the intra-similariy is smaller and the inter-similarity is higher in the $\textit{DGoW}$ graph construction. Therefore, we can better separate between text structures in different classes using the $\textit{DGoW}$ configuration. %As we can see, the predictions values illustrate that the class probabilities don't usually add up to 1.

\begin{table*}[t]
\centering
% \resizebox{1.5\columnwidth}{!}{%
% \scriptsize
\caption{Structural similarity between different pairs of
labels for OH dataset\textcircled{1} MGoW Configuration \textcircled{2} DGoW Configuration.}\label{tab:OHsim}

\begin{tabular}{lr|rrrrrr} \toprule

& \textbf{Labels}  & $\mathbf{\boldsymbol{\omega}=2}$ & $\mathbf{\boldsymbol{\omega}=5}$  & $\mathbf{\boldsymbol{\omega}=10}$  & $\mathbf{\boldsymbol{\omega}=15}$& $\mathbf{\boldsymbol{\omega}=20}$\\ \cmidrule{1-7}
\multirow{10}{*}{\textcircled{1}} 
&\textit{C04/C04}  & 64.08 & 54.58 & 51.88 & 51.01 & 50.63 \\
&\textit{C10/C10}  & 65.46 & 52.31 & 49.32 & 48.01 & 46.17 \\ 
&\textit{C14/C14}  & 67.35 & 56.61 & 51.60 & 48.69 &  50.24 \\ 
&\textit{C23/C23}  & 60.31 & 48.10 & 44.39 & 42.42 & 41.70 \\
&\textit{C04/C10}  & 59.80 & 44.79 & 41.04 & 39.22  & 38.26\\ 
&\textit{C04/C14}  & 56.76 & 42.12 & 36.76 & 34.58 &  34.95\\ 
&\textit{C04/C23}  & 58.93 & 45.55 &  41.65 & 39.78 & 39.37  \\
&\textit{C10/C14}  & 62.06 & 46.66 & 41.70 & 39.67 &  39.20\\ 
&\textit{C10/C23}  & 61.31 & 47.55 & 43.85 & 42.09 & 40.88 \\
&\textit{C14/C23}  & 61.72 & 48.17 & 43.33 & 41.02 & 40.86 \\ \cmidrule{1-7}

\multirow{10}{*}{\textcircled{2}} 
&\textit{C04/C04}  & 83.95 &  73.89 & 71.44 & 71.49 & 71.26\\
&\textit{C10/C10}  & 88.07 &  80.25 & 77.65 & 77.96 & 78.11\\ 
&\textit{C14/C14}  & 82.41 &  71.11 & 69.61  & 67.89 & 67.86 \\ 
&\textit{C23/C23}  & 84.27 &  73.80 & 70.53 & 70.15 &  70.16\\
&\textit{C04/C10}  & 0     & 0      & 0&0 &0 \\ 
&\textit{C04/C14}  & 0     & 0      &  0&0& 0 \\ 
&\textit{C04/C23}  & 0     &  0 & 0  & 0& 0 \\
&\textit{C10/C14}  & 0     &  0 &  0&0 &0  \\ 
&\textit{C10/C23}  & 0     &  0 & 0 & 0&  0\\
&\textit{C14/C23}  & 0     &  0 & 0 &0 & 0 \\
\midrule
\end{tabular}
% }
% \captionsetup{width=1.5\columnwidth}
\end{table*}

\subsection{Structural Similarity Using FastGAE Embedding}
\label{sec:appendix_B}

\begin{table*}[htbp]
\centering
% \resizebox{1.5\columnwidth}{!}{%
% \scriptsize
\caption{Structural similarity between different pairs of labels for R8 dataset \textcircled{1} Configuration MGoW \textcircled{2} Configuration DGoW.}\label{tab:R8Fats}
\begin{tabular}{lr|rrrrrr} \toprule
&  \textbf{Labels}  & $\mathbf{\boldsymbol{\omega}=2}$ & $\mathbf{\boldsymbol{\omega}=5}$  & $\mathbf{\boldsymbol{\omega}=10}$  & $\mathbf{\boldsymbol{\omega}=15}$& $\mathbf{\boldsymbol{\omega}=20}$\\ \cmidrule{1-7}
\multirow{3}{*}{\textcircled{1}} 
&\textit{earn/earn} & 61.80 & 63.99 & 61.57 & 55.52 & 61.10 \\
&\textit{acq/acq}   & 66.70 & 83.16 & 87.22 & 87.36 & 9.89 \\
&\textit{earn/acq}  & -3.02 & -4.60 & -8.21 & -2.36 & -15.20 \\ \cmidrule{1-7}
\multirow{3}{*}{\textcircled{2}} 
&\textit{earn/earn} & 99.81 &  98.91 & 99.93  &  99.99 & 99.87 \\
&\textit{acq/acq}   & 99.86 &  99.98 & 99.99  &  99.99 & 99.99  \\
&\textit{earn/acq}  &-38.07 & -32.61 & -34.70 & -64.44 & -32.89 \\
\midrule
\end{tabular}
% }
% \captionsetup{width=1.5\columnwidth}
\end{table*}

To further support the spectral embedding results, we use the FastGAE model to generate the structural embeddings. We report the result of the similarity comparison for OH and R8 datasets in Tables \ref{tab:R8Fats} and \ref{tab:OHFats}. We notice the same trends as spectral embeddings. The intra-similarity is always high and positive to the extent that some sentences in some classes exhibit greater structural similarity to a different class than the sentences in the same class. The inter-similarity is much higher and almost equal to 1 in the DGoW configuration. The intra-similarity is very small in  DGoW compared to MGoW values. Some values are usually negative, which indicate that the graph representations of nodes and sentences in different classes are strongly opposite vectors.

% In Tables \ref{tab:R8Fats} and \ref{tab:OHFats}, we report the result of the experiment comparing the \textit{intra-similarity} and \textit{inter-similarity} using \textit{FastGAE} embeddings for the two dataset \textbf{R8} and \textbf{OH}. 

\begin{table*}[t]
\centering
% \resizebox{1.5\columnwidth}{!}{%
% \scriptsize
\caption{Structural similarity between different pairs of labels for OH dataset \textcircled{1} Configuration MGoW \textcircled{2} Configuration DGoW.}\label{tab:OHFats}

\begin{tabular}{lr|rrrrrr} \toprule
&  \textbf{Labels}  & $\mathbf{\boldsymbol{\omega}=2}$ & $\mathbf{\boldsymbol{\omega}=5}$  & $\mathbf{\boldsymbol{\omega}=10}$  & $\mathbf{\boldsymbol{\omega}=15}$& $\mathbf{\boldsymbol{\omega}=20}$\\ \cmidrule{1-7}
\multirow{20}{*}{\textcircled{1}} 
&\textit{C23/C23} & 47.44 & 57.35 & 37.23 & 38.00 & 37.62 \\ 
&\textit{C10/C10} & 56.38 & 79.58 & 63.42 & 49.23 & 47.63 \\ 
&\textit{C04/C04} & 47.56 & 72.69 & 51.70 & 39.83 & 43.31 \\ 
&\textit{C14/C14} & 48.54 & 38.79 & 10.91 & 51.74 & 55.45 \\ 
&\textit{C20/C20} & 50.56 & 66.71 & 44.35 & 41.90 & 42.47 \\ 
&\textit{C21/C21} & 58.43 & 78.70 & 58.09 & 55.21 & 54.51 \\ 
&\textit{C10/C23} & 49.51 & 67.53 & 48.66 & 41.62 & 41.02 \\ 
&\textit{C04/C23} & 45.56 & 64.63 & 44.02 & 32.57 & 33.77 \\ 
&\textit{C14/C23} & 47.90 & 47.22 & 20.10 & 42.17 & 41.0 \\ 
&\textit{C20/C23} & 46.56 & 61.91 & 40.76 & 28.66 & 29.23 \\ 
&\textit{C21/C23} & 49.51 & 67.12 & 46.62 & 40.31 & 40.37 \\ 
&\textit{C04/C10} & 47.69 & 76.09 & 57.32 & 39.09 & 40.21 \\ 
&\textit{C10/C14} & 49.06 & 55.32 & 25.92 & 42.50 & 40.51 \\ 
&\textit{C10/C20} & 48.31 & 72.91 & 53.15 & 32.83 & 31.63 \\ 
&\textit{C10/C21} & 55.06 & 79.19 & 60.79 & 49.75 & 48.56 \\ 
&\textit{C04/C14} & 44.71 & 53.03 & 23.62 & 28.92 & 25.51 \\ 
&\textit{C04/C20} & 43.34 & 69.71 & 48.00 & 38.96 & 39.37 \\ 
&\textit{C04/C21} & 47.28 & 75.65 & 54.90 & 38.99 & 41.99 \\ 
&\textit{C14/C20} & 47.10 & 50.78 & 21.77 & 23.93 & 21.26 \\ 
&\textit{C14/C21} & 49.52 & 54.96 & 24.90 & 41.12 & 39.72 \\ 
&\textit{C20/C21} & 43.86 & 72.49 & 50.88 & 29.81 & 30.05 \\ 
  \cmidrule{1-7}

\multirow{20}{*}{\textcircled{2}} 
&\textit{C23/C23} & 100.0 & 100.0 & 99.98 & 100.0 & 99.99 \\ 
&\textit{C10/C10} & 100.0 & 100.0 & 99.95 & 100.0 & 99.99 \\ 
&\textit{C04/C04} & 100.0 & 100.0 & 99.98 & 100.0 & 99.99 \\ 
&\textit{C14/C14} & 100.0 & 100.0 & 99.98 & 100.0 & 99.99 \\ 
&\textit{C20/C20} & 100.0 & 100.0 & 99.99 & 100.0 & 99.99 \\ 
&\textit{C21/C21} & 100.0 & 100.0 & 99.99 & 100.0 & 99.99 \\ 
&\textit{C10/C23} & -9.52 & -10.08 & -6.09 & -9.96 & -2.98 \\ 
&\textit{C04/C23} & -13.04 & -15.41 & -13.61 & -12.22 & -10.2 \\ 
&\textit{C14/C23} & -13.96 & -13.63 & -13.19 & -12.89 & -7.23 \\ 
&\textit{C20/C23} & -12.58 & -9.32 & -6.7 & -7.93 & -11.44 \\ 
&\textit{C21/C23} & -9.63 & -10.31 & -9.61 & -9.02 & -8.03 \\ 
&\textit{C04/C10} & -11.53 & -9.97 & -7.37 & -9.9 & -3.45 \\ 
&\textit{C10/C14} & -10.84 & -10.69 & -7.53 & -9.44 & -5.05 \\ 
&\textit{C10/C20} & -10.17 & -7.72 & -2.39 & -7.5 & -7.29 \\ 
&\textit{C10/C21} & -7.32 & -7.51 & -3.36 & -6.61 & 1.28 \\ 
&\textit{C04/C14} & -13.76 & -13.78 & -15.0 & -12.93 & -12.42 \\ 
&\textit{C04/C20} & -13.19 & -10.03 & -6.88 & -9.81 & -12.27 \\ 
&\textit{C04/C21} & -9.82 & -10.93 & -9.13 & -10.56 & -8.55 \\ 
&\textit{C14/C20} & -14.3 & -9.53 & -10.82 & -8.57 & -13.27 \\ 
&\textit{C14/C21} & -10.87 & -11.64 & -11.88 & -9.37 & -2.21 \\ 
&\textit{C20/C21} & -15.64 & -8.61 & -2.39 & -8.11 & -7.15 \\ 

\midrule
\end{tabular}
% }
% \captionsetup{width=1.5\columnwidth}
\end{table*}

\section{Examples of Predictions }
\label{sec:appendix_C}

In Table \ref{tab:examples}, we give the predictions of \textit{DGoW-GNN} for some randomly selected test sentences of the R8 dataset.
\begin{table*}[h]
    \centering
\caption{Five randomly sampled examples of \textit{DGoW-GNN} predictions.}
\label{tab:examples}
\resizebox{\columnwidth}{!}{
\begin{tabular}{ c|l } 
\hline
\textbf{ Ground Truth} & \textbf{Test Sentence and Predictions} \\ \hline  \\
earn &  \begin{minipage}{15cm} \textbf{Sentence : } \textit{entre computer centers inc etre nd qtr loss shr loss cts vs  profit cts net loss vs profit revs mln vs mln st half shr loss cts vs profit cts net loss vs profit revs mln vs mln note current year net both periods includes dlr pretax provision for closing overseas operations and tax credits dlrs in quarter and dlrs in half reuter } \\
 \\ \textbf{Predictions : } acq: $9.68 ~10^{-9}$, crude: $4.76 ~10^{-9}$, \textbf{earn: $\mathbf{0.99}$}, grain: $1.97 ~10^{-8}$, interest: $7.40 ~10^{-9}$, money-fx: $1.45 ~10^{-8}$, ship: $1.16 ~10^{-08}$, trade: $8.33 ~10^{-9}$ 
 \\ 
 \end{minipage}  \\  \hline  \\

 earn &  \begin{minipage}{15cm} \textbf{Sentence : } \textit{weirton steel corp rd qtr net mln vs mln revs mln vs mln nine mths net mln vs mln revs mln vs mln note company does not report per share earnings as it is a privately owned concern net amounts reported are before taxes profit sharing and contribution to employee stock ownership trust reuter  } \\
 \\ \textbf{Predictions : } acq: $6.56 ~10^{-8}$, crude: $2.44 ~10^{-8}$, \textbf{earn: $\mathbf{0.99}$}, grain: $1.18 ~10^{-8}$, interest: $1.08 ~10^{-8}$, money-fx: $1.19 ~10^{-8}$, ship: $1.64~10^{-8}$, trade: $6.27 ~10^{-9}$ 
 \\ 
 \end{minipage}   \\  \hline  \\

  acq &  \begin{minipage}{15cm} \textbf{Sentence : } \textit{lvi group lvi to make acquisition lvi group inc said it has agreed in principle to purchase all outstanding shares of spectrum holding corp for a proposed mln dlrs in cash lvi said an additional mln dlrs in common stock and seven mln dlrs in notes will become payable if spectrum has certain minimum future earnings lvi an interior construction firm said the acquisition is subject to execution of a definitive agreement and completion of due diligence lvi and spectrum an asbestos abatement concern expect to close the deal in june lvi said reuter  } \\
 \\ \textbf{Predictions : } \textbf{acq: $\mathbf{5.55 ~10^{-2}}$}, crude: $3.53 ~10^{-8}$, earn: $1.26 ~10^{-4}$, grain: $2.76 ~10^{-8}$, interest: $1.28 ~10^{-8}$, money-fx: $9.17 ~10^{-9}$, ship: $1.24 ~10^{-7}$, trade: $8.46 ~10^{-8}$ 
 \\ 
 \end{minipage}  \\   \hline \\

   interest &  \begin{minipage}{15cm} \textbf{Sentence : } \textit{average yen cd rates fall in latest week average interest rates on yen certificates of deposit cd fell to pct in the week ended april from pct the previous week the bank of japan said new rates previous in brackets average cd rates all banks pct money market certificate mmc ceiling rates for week starting from april pct average cd rates of city trust and long term banks less than days pct days pct average cd rates of city trust and long term banks days pct days pct days unquoted days pct over days pct unqtd average yen bankers acceptance rates of city trust and long term banks to less than days pct days pct days unquoted unqtd reuter   } \\
 \\ \textbf{Predictions : } acq: $4.33 ~10^{-4}$, crude: $2.01 ~10^{-7}$, earn: $1.30 ~10^{-7}$, grain: $2.71 ~10^{-7}$, \textbf{interest:} $\mathbf{0.99}$, money-fx: $2.60 ~10^{-4}$, ship: $1.81 ~10^{-6}$, trade: $5.22 ~10^{-5}$ 
 \\ 
 \end{minipage}  \\   \hline \\

   trade &  \begin{minipage}{15cm} \textbf{Sentence : } \textit{white house says japanese tarriffs likely the white house said high u s tariffs on japanese electronic goods would likely be imposed as scheduled on april despite an all out effort by japan to avoid them presidential spokesman marlin fitzwater made the remark one day before u s and japanese officials are to meet under the emergency provisions of a july semiconductor pact to discuss trade and the punitive tariffs fitzwater said i would say japan is applying the full court press they certainly are putting both feet forward in terms of explaining their position but he added that all indications are they the tariffs will take effect reuter  } \\
 \\ \textbf{Predictions : } acq: $7.00 ~10^{-8}$, crude: $6.74 ~10^{-7}$, earn: $8.41 ~10^{-9}$, grain: $2.53 ~10^{-5}$, interest: $5.30 ~10^{-6}$, money-fx: $9.28 ~10^{-4}$, ship: $3.95~10^{-7}$, \textbf{trade: }$\mathbf{0.99}$ 
 \\ 
 \end{minipage}  \\
 \hline
\end{tabular}}
\end{table*}

\section{Ablation Studies}

In this appendix we provide extensive ablation study results.

\subsection{Window Size} \label{sec:WindowSizeAblation}

In Table \ref{tab:window}, we compare the performance of DGoW on the datasets R8, OH and MR when using different window sizes 2, 5, 10 and 15.

\begin{table}[ht]\centering
% \tiny
\caption{Ablation study on the window size $\omega.$}\label{tab:window}
% \scriptsize
\begin{tabular}{llll}\toprule
\textbf{Model}   & \textbf{R8}  & \textbf{OH}  & \textbf{MR} \\ \cmidrule{1-4}
DGoW w/ $\omega=2$ & \textbf{\textcolor{black}{95.17 (0.22)} }& \textbf{\textcolor{black}{44.59 (1.01)}} & \textbf{\textcolor{black}{71.65 (0.39)} }\\
DGoW w/   $\omega=5$ & 82.64 (0.19)   & 43.33 (1.94)  & 61.47 (0.40) \\
DGoW w/ $\omega=10$  & 73.37 (0.17)   & 41.21 (0.95) &  62.02 (0.66) \\
\bottomrule
\end{tabular}
\end{table}

We notice that the accuracy is decreasing when increasing the window size.

\subsection{Word Embedding} \label{sec:WordEmbeddingAblation}
In Table \ref{tab:ablation_embedding}, we compare the performance of DGoW on the datasets R8, OH and MR when using the one-hot embeddings and the pre-trained GLoVe embeddings of dimension 100.

\begin{table}[ht]\centering
\caption{Ablation study on the word embedding }\label{tab:ablation_embedding}
% \scriptsize
\begin{tabular}{llll}\toprule
\textbf{Model}   & \textbf{R8}  & \textbf{OH}  & \textbf{MR} \\ \cmidrule{1-4}

DGoW w/ One-Hot & \textbf{\textcolor{black}{95.17 (0.22)}} & \textbf{\textcolor{black}{44.59 (1.01)} } & \textbf{\textcolor{black}{71.65 (0.39)}} \\
DGoW w/ GloVe   & 80.49 (0.04) &  41.44 (0.84) &   62.32 (0.67)\\
\bottomrule
\end{tabular}
\end{table}

\subsection{GNN Model} \label{sec:GNNAblation}
In Table \ref{tab:ablation_gnn}, we compare the performance of DGoW on the datasets R8,  OH and MR when using different GNN architectures : GCN and GAT \cite{veličković2018graph}.

\begin{table}[ht]\centering
\caption{Ablation study on the GNN architecture}\label{tab:ablation_gnn}
% \scriptsize
\begin{tabular}{llll}\toprule
\textbf{Model}   & \textbf{R8} & \textbf{OH}  & \textbf{MR} \\ \cmidrule{1-4}
DGoW w/ GCN & \textbf{\textcolor{black}{95.17 (0.22)}} &  \textbf{\textcolor{black}{44.59 (1.01)}} & \textbf{\textcolor{black}{71.65 (0.39)} }\\
DGoW w/ GAT   & 79.84 (3.07) &  11.50 (1.20)  &  62.43 (0.86)  \\
\bottomrule
\end{tabular}
\end{table}

\subsection{Aggregation Function} \label{sec:AggregatorAblation}

\begin{table}[h]\centering
% \tiny
% \scriptsize
\caption{Ablation study on the aggregation function.} \label{tab:ablation}
\begin{tabular}{lllll}\toprule
\textbf{Aggregator } & \textbf{R8} & \textbf{R52}  & \textbf{OH}  & \textbf{MR} \\ \midrule

DGoW w/ $\text{\small AVG}$ & \textbf{\textcolor{black}{95.17 (0.22)}} & \textbf{\textcolor{black}{86.26 (1.54)}} & \textbf{\textcolor{black}{44.59 (1.01)}} & \textbf{\textcolor{black}{71.65 (0.39)}} \\
DGoW w/ $\text{\small MLP}$ &  \textcolor{black}{92.16 (1.38)} & \textcolor{black}{82.08 (1.95)} & \textcolor{black}{ 29.79 (5.13)} & \textcolor{black}{70.78 (0.87)}  \\
DGoW w/ $\text{\small PROD}$ &  \textcolor{black}{90.22 (5.30)} & \textcolor{black}{74.13  (1.50)} & \textcolor{black}{ 25.01 (25.39)} & \textcolor{black}{69.09 (0.58)}  \\
\bottomrule
\end{tabular}
\end{table}

We first tested the $\text{\small MLP}$ aggregator that take the concatenation of the first and last element of the Bi-LSTM and feed it to an MLP. We also tested $\text{\small PROD}$ function defined as follows
\begin{equation*}
    % \text{\small PROD} \left( \left [  \Tilde{h}^{(p)}_{w^{(s)}_1},\ldots,\Tilde{h}^{(p)}_{w^{(s)}_{L(s)}}\right ] \right ) = 
    \prod_{(i,j)\in\mathcal{W}} \sigma\left( \Tilde{h}^{(p)}_{w^{(s)}_i} {}^{T}  ~~\Tilde{h}^{(p)}_{w^{(s)}_{j}}\right), 
\end{equation*}
where is $\mathcal{W}$ is the set of all the windows of size $w$. By using the  $\text{\small PROD}$ function, we  assume that a sentence belongs to a disconnected subgraph if only if every edge belongs to that disconnected subgraph. In Table \ref{tab:ablation}, we observe the impact of the $\text{\small AVG},$ $\text{\small MLP}$ and $\text{\small PROD}$  aggregation function on the performance of our DGoW-GNN on all our considered datasets. As noticed, we obtain the best results using the $\text{\small AVG}$ aggregator. It is also the aggregation with the smallest standard deviation.

%%%%%%%%%%%%%%%%%%%%%%%%%%%%%%%%%%%%%%%%%%%%%%%%%%%%%%%%%%%%

\end{document}